\documentclass[11pt,a4paper]{article}
\usepackage[hyperref]{acl2017}
\usepackage{times}
\usepackage{latexsym}
\usepackage{url}

\usepackage{caption}
\usepackage{subcaption}
\usepackage{tabularx}
\usepackage{booktabs}
\usepackage{multirow}
\usepackage{graphicx}
\usepackage{amsmath}
\usepackage{comment}
\usepackage{xcolor,colortbl}
\usepackage{enumitem}

\aclfinalcopy 

\title{Inference of Fine-Grained Event Causality from Blogs and Films}

\author{Zhichao Hu, Elahe Rahimtoroghi and Marilyn A Walker \\
  Natural Language and Dialogue Systems Lab \\
  Department of Computer Science, University of California Santa Cruz \\
  Santa Cruz, CA 95064, USA \\
  {\tt zhu@soe.ucsc.edu},  {\tt elahe@soe.ucsc.edu}, {\tt mawalker@ucsc.edu}\\}

\date{}

\begin{document}
\maketitle
\begin{abstract}
Human understanding of narrative is mainly driven by reasoning about
causal relations between events and thus recognizing them is a key capability for computational models of
language understanding. 
Computational work in this area has approached
this via two different routes: by focusing on acquiring a knowledge
base of common causal relations between events, or by attempting to
understand a particular story or macro-event, along with its
storyline.  
In this position
paper, we focus on knowledge acquisition approach and claim that newswire is a relatively poor source for learning
fine-grained causal relations between everyday events.  
We describe
experiments using an unsupervised method to learn causal relations
between events in the narrative genres of first-person narratives and
film scene descriptions. We show that our method learns fine-grained
causal relations, judged by humans as likely to be
causal over 80\% of the time. We also demonstrate that the learned
event pairs do not exist in publicly available event-pair datasets
extracted from newswire.
\end{abstract}

\section{Introduction}
\label{sec:intro}

Computational models of language understanding must recognize
narrative structure because many types of natural language texts are
narratively structured, e.g.  news, reviews, film scripts,
conversations, and personal blogs
\cite{Polanyi89,Jurafskyetal14,bell2005news,Gordonetal11}.  Human
understanding of narrative is driven by reasoning about causal
relations between the events and states in the story
~\cite{Gerrig93,Graesseretal94,Lehnert81,Goyaletal10}.  Thus previous
work has aimed to learn a knowledge base of semantic relations between
events from text
\cite{ChklovskiPantel04,Gordonetal11,ChambersJurafsky08,Balasubramanianetal13,PichottaMooney14,DoChaRo11},
with the long-term aim of using this knowledge for understanding.
Some of this work explicitly models causality; other work
characterizes the semantic relations more loosely as ``events that
tend to co-occur''.  Related work points out that causality is
granular in nature, and that humans flexibly move back and forth
between different levels of granularity of causal knowledge
\cite{Hobbs85}.  Thus methods are needed to learn causal
relations and reason about them at different levels of granularity
~\cite{Mulkar2011}.

\begin{figure}[t]
\small
\begin{tabularx}{\columnwidth}{X}
\toprule
{\bf We packed all our things} on the night before Thu (24 Jul) except for frozen food. We brought a lot of things along. {\bf We woke up} early on Thu and JS started {\bf packing the frozen marinatinated food} inside the small cooler... In the end, we decided the best place to set up the tent was the squarish ground that's located on the right. Prior to setting up our tent, {\bf we placed a tarp on the ground}. In this way, the underneaths of the tent would be kept clean. After that, {\bf we set the tent up}.  \\
\bottomrule
\end{tabularx}
\caption{Part of a blog story about camping}
\label{fig:camping}
\end{figure}

One limitation of prior work is that it has primarily focused on
newswire, 
thus have only learned relations about newsworthy
topics, and likely the
most frequent, highly common (coarse-grained)
news events.
But news articles are not the only resource for learning
about relations between events.  Much of the content on social media
in personal blogs is written by ordinary people about their daily
lives \cite{Burtonetal09}, and these blogs contain a large variety of
everyday events \cite{Gordonetal12}. Film scene descriptions are also
action-rich and told in fine-grained detail
\cite{BeamerGirju09,Huetal13}.  Moreover, both of these genres
typically report events in temporal order, which is a primary cue to
causality. In this position paper, we claim that knowledge about
fine-grained causal relations between everyday events is often not
available in news, and can be better learned from other narrative
genres.

For example, Figure~\ref{fig:camping} shows a part of a personal
narrative written in a blog about a camping trip
\cite{Burtonetal09}. The major event in this story is
\textit{camping}, which is contingent upon several finer-grained
events, such as \textit{packing things the night before, waking up in
  the morning, packing frozen food}, and later on at the campground,
\textit{placing a tarp} and \textit{setting up the tent}.  Similarly
film scene descriptions, such as the one shown in
Figure~\ref{lotr-fig}, typically contain fine-grained causality. In
this scene from Lord of the Rings, \textit{grabbing} leads to
\textit{spilling}, and {\it pushing} leads to {\it stumbling} and {\it
  falling}.

We show that unsupervised methods for modeling causality can learn
fine-grained event relations from personal narratives and film scenes,
even when the corpus is relatively small compared to those that have
been used for newswire.  We learn high-quality
causal relations, with over 80\% judged as causal by humans. We
claim that these fine-grained causal relations are much closer in
spirit to those motivating earlier work on scripts~\cite{Lehnert81,
  Schank77, wilensky82, deJong79}, and we show that the causal
knowledge we learn is not found in causal knowledge bases learned from
news.

Section~\ref{sec:background} first summarizes previous work on
learning causal knowledge. We then present our experiments and results
on modeling event causality in blogs and film scenes in
Section~\ref{sec:method}.  Conclusions and future directions are discussed in
Section~\ref{sec:conclusion}.

\section{Background and Related Work}
\label{sec:background}

Cognitive theories of narrative understanding define 
narrative coherence in terms of four different
sources of causal inferences between events A and B
\cite{trabassoVandenBroek85,warren1979event,trabasso1989logical,van1990causal}.  (1)
Physical: A physically causes event B.  (2) Motivational: A happens
with B as a motivation.  (3) Psychological: A brings about emotions
expressed by event B.  (4) Enabling: A creates a state or condition
for B to happen.  

\begin{figure}[t]
\small
\begin{tabularx}{\columnwidth}{X}
\toprule
Pippin, sitting at the bar, chatting with
Locals. Frodo leaps to his feet and pushes his way towards the
bar. Frodo \textbf{grabs} Pippin's sleeve, \textbf{spilling} his
beer. Pippin \textbf{pushes} Frodo away...he \textbf{stumbles}
backwards, and \textbf{falls} to the floor. \\ 
\bottomrule
\end{tabularx}
\caption{Film Scene from Lord of the Rings, Fantasy Genre}
\label{lotr-fig}
\end{figure}

There has been a great deal of interest in learning narrative
relations or narrative schema in an unsupervised or weakly
supervised manner from text. Here we focus on
work where the resulting knowledge bases have been made publicly
available, allowing us to compare the learned knowledge directly.

The VerbOcean project learned five different semantic relations
between event types (verbs) from newswire, with the {\sc
  happens-before} relation defined as ``indicating that the two verbs
refer to two temporally disjoint intervals or instances''.  WordNet's
cause relation, between a causative and a resultative verb (as in
buy::own) is tagged as an instance of {\sc happens-before} in
VerbOcean, consistent with the heuristic that temporal ordering is a
major component of causality.  Other examples of the {\sc
  happens-before} relation in the VerbOcean knowledge base include
marry::divorce, detain::prosecute, enroll::graduate,
schedule::reschedule, and tie::untie \cite{ChklovskiPantel04}.

\newcite{Balasubramanianetal13} generate pairs of event relational
tuples, called \textit{Rel-grams}. The Rel-grams are publicly
available through an online search
interface\footnote{http://relgrams.cs.washington.edu:10000/relgrams}.
Rel-gram tuples are extracted using a co-occurrence statistical
metric, Symmetric Conditional Probability (SCP), which combines Bigram
probability in both directions as follows:
\begin{equation}
SCP(e_1, e_2) = P(e_2|e_1) \times P(e_1|e_2)
\end{equation}

Their evaluation experiments directly compared the knowledge learned
in Rel-grams to the previous work on narrative schemas \cite{ChambersJurafsky08,ChambersJurafsky09},
showing that they achieve better results, thus our work compares
directly to the tuples available in Rel-grams.

Other work focuses more directly on learning {\bf causal}
or {\bf contingency} relations between
events. \newcite{BeamerGirju09} introduced a distributional measure
called \textit{Causal Potential} to assess the likelihood of 
a causal relation holding between two events. This measure
is based on Suppes' probabilistic theory of causality 
\cite{suppes1970probabilistic}.

\begin{equation}
\mathit{CP}(e_1, e_2) = \mathit{PMI}(e_1, e_2) + \log\frac{P(e_1\rightarrow e_2)}{P(e_2\rightarrow e_1)}\\
\label{eq:cp}
\end{equation}
\begin{equation*}
\mathrm{where}~\mathit{PMI}(e_1, e_2) = \log\frac{P(e_1, e_2)}{P(e_1)P(e_2)}
\end{equation*}
 
\noindent where the arrow notation means ordered event pairs, i.e. event $e_1$
occurs before event $e_2$. CP consists of two terms: the first is
pair-wise mutual information (PMI) and the second is relative ordering
of bigrams. PMI measures how often events occur as a pair (without
considering their order); whereas relative ordering accounts for the
order of the event pairs because temporal order is one of the
strongest cues to causality \cite{BeamerGirju09,RiazGirju10,riaz2013toward}.
This work explicitly links their definitions
to research using the Penn Discourse
Treebank (PDTB) definition of {\sc contingency}.

\newcite{BeamerGirju09} applied the CP measure to 173 film scripts,
resulting in a high
correlation between human-judged causality and the CP measure.  Their
paper provides a list of 90 verb pairs, selected from the high,
middle and low CP ranges in their learned causal pairs.  We compare
their 30 highest CP events with causal event pairs that we learn from
film. 

\newcite{RiazGirju10} apply a similar measure to topic-sorted news
stories about Hurricane Katrina and the Iraq War and present ranked
causality relations between events for these topics, suggesting that
topic-sorted corpora can produce better causal knowledge.  Other work
has also used CP to measure the contingency relation between two
events, reporting better results than achieved with PMI or bigrams
alone \cite{Huetal13,Rahimtoroghietal16}.

\section{Methods and Evaluations}
\label{sec:method}

Our primary goal is simply to show that fine-grained causal relations
can be learned from film scripts and blogs, and that these are not
found in causal knowledge bases learned from newswire.  In this section we describe our datasets and methods, and the present two evaluations.
First, we evaluate whether the relations learned are causal using human judgment HITs on Amazon Mechanical Turk. Second, we directly compare to event pair collections from other publicly available
sources learned from news genre.

\begin{table}[t]
\centering
\small
\begin{tabularx}{2.2in}{ c c c }
\toprule
{\bf Corpus} & {\bf Number} & {\bf Word Count} \\
\midrule
Drama & 579 & 6,680,749 \\
Fantasy & 113 & 1,186,587 \\
Mystery & 107 & 1,346,496 \\
Camping & 1,062 & 2,207,458 \\
\bottomrule
\end{tabularx}
\caption{Number of documents and word count for each dataset \label{tab:dataset}}
\end{table}

\subsection{Datasets}
Topical coherence and similarity of events within the corpus used for
learning event relations can be as important as the size of the
corpus~\cite{RiazGirju10, Rahimtoroghietal16}.  
We use two datasets for learning causal event pairs: first-person 
narratives from blogs~\cite{Burtonetal09, Rahimtoroghietal16}, 
and film scene descriptions (excluding dialogs because dialogs 
are not as action-rich)~\cite{Walkeretal12d, Huetal13}. 
Our experiment
on blogs learns causal relations from a
topic-sorted corpus of $\sim$1000 camping stories.
We also posit that the genre
of a film may select for similar types of events. However
genres can be defined broadly or narrowly, 
e.g. the Drama genre overlaps with many other genres. 
We thus compare two narrow film genres of
Fantasy and Mystery with the Drama genre from an existing
corpus~\cite{Walkeretal12d, Huetal13}.  The raw numbers for each
subcorpus are shown in Table~\ref{tab:dataset}.  
Note that Camping corpus consists of blog posts which
are much shorter compared to movie scripts. Thus their 
word count is much smaller compared to films corpus 
despite the larger number of documents.

\subsection{Methods}

In the blogs,
related event pairs are more frequently separated by utterances that
provide state descriptions or affective reactions to events
\cite{Swansonetal14b}. As a result, we use Causal Potential (CP) 
measure to assess the causal relation between events and apply 
skip-2 bigram method for modeling event pairs. 
But in film scenes, events are very densely distributed, thus related
event pairs are often adjacent to one another and therefore nearby 
events are more likely to be causal. So, for event pairs extracted 
from films we use a variant of CP measure, shown in Eq.~\ref{eq:cpc}, that
accounts for
different window sizes and punishes event pairs from larger window
sizes~\cite{RiazGirju10,riaz2013toward,DoChaRo11,PichottaMooney14}. 

\begin{equation}
\label{eq:cpc}
CP_{variant} (e_1, e_2) = \sum_{i = 1}^{w_{max}} \frac{CP_i (e_1, e_2)}{i}
\end{equation}

\noindent where $w_{max}$ is the max window size (how many
events after the current event are paired with the
current event). $CP_i(e_1; e_2)$ is the CP score for
event pair $e_1; e_2$ calculated using window size $i$.

\begin{table}
\small
\centering
\begin{tabularx}{2.75in}{X}
\toprule
{\bf Camping Event Pairs} \\
\midrule
person - pack up $\rightarrow$ person - go - home  \\
person - wake up $\rightarrow$ person - pack up - backpack \\
person - eat - breakfast $\rightarrow$ person - pack up - campsite  \\
person - head $\rightarrow$ hike up  \\
person - pack up - car $\rightarrow$ head out \\
\midrule
{\bf Fantasy Event Pairs} \\ 
\midrule
person - slam - something $\rightarrow$ shut  \\
send - something $\rightarrow$ fly - something \\
person - watch $\rightarrow$ something - disappear  \\
person - pick up - something $\rightarrow$ carry - something \\
person - turn $\rightarrow$ face - person  \\
\midrule
{\bf Mystery Event Pairs}  \\
\midrule
 bind  $\rightarrow$ gag  \\
 person - reach  $\rightarrow$ touch - something \\
 person - pull - something $\rightarrow$ reveal - something \\
 person - look  $\rightarrow$ confuse \\
 person - come $\rightarrow$ rest \\
 \midrule
{\bf Drama Event Pairs}  \\
\midrule
 person - slam - something  $\rightarrow$ shut  \\
 person - offer - something  $\rightarrow$ something - decline \\
 person - rummage $\rightarrow$ person - find - something \\
 send - something  $\rightarrow$ something - fly \\
 send - something  $\rightarrow$ sprawl \\
\bottomrule
\end{tabularx}
\caption{High-CP pairs from Camping, Fantasy and Mystery datasets}
\label{tab:cp-pairs}
\end{table}


\subsection{Experiments and Results}

We process the data in each dataset and calculate causal 
potential score for each extracted event pair, resulting in a 
rank-ordered list of causal event pairs. 
We evaluate the top 100 event pairs for camping, and the top 684 
event pairs for films. We take a number of event pairs from each 
film genre (proportional to the number of films in that genre, see 
Table~\ref{tab:dataset} and~\ref{tab:amt-results}), then remove
duplicate event pairs, which result in the 684 event pairs from film.
Table~\ref{tab:cp-pairs} presents examples of learned 
high-CP event pairs from each corpus.
In our following Mechanical Turk experiments, Turkers have to pass 
qualification tests similar to the actual HITs to be able to participate 
in our task.

\begin{table}[t]
\centering
\small
\begin{tabularx}{2.6in}{ c c c }
\toprule
{\bf Genre} & {\bf \# High-CP Pairs} & {\bf \% Causality} \\
\midrule
Drama & 655 & 82.6 \\
Fantasy & 127 & 90.7 \\
Mystery & 122 & 87.7 \\
\bottomrule
\end{tabularx}
\caption{Percentage of high-CP pairs labeled as causal by AMT worker, comparing with low-PC pairs, in film genres Drama, Fantasy and Mystery.}
\label{tab:amt-results}
\end{table}

In a study on each genre of films, we compare high-CP pairs to 
a random sample of low-CP pairs on Mechanical Turk to 
see if pairs with high CP score more strongly encode causal 
relations that ones with low CP. 
For every event pair in the 684 high pairs, we randomly select a low
pair in order to collect human judgments on Mechanical Turk. The task
first defines events and event pairs, then gives examples of event pairs
with causal relations. Turkers are asked to select the event pair that 
is more likely to manifest a causal relation. 
The results, summarized in 
Table~\ref{tab:amt-results}, show that humans judge a large 
majority of the high-CP pairs to have a causal relation and the 
results vary by genre. The causality rate is achieved for more 
focused genres, Fantasy (90.7\%) and Mystery (87.7\%), despite their
smaller size, and the lowest for Drama (82.6\%).  We believe this result is
further evidence that topical coherence improves causal relation
learning~\cite{Rahimtoroghietal16, RiazGirju10}.

In our second evaluation method, we compare the learned CP 
event pairs to the existing causal knowledge collections.
First, we compare our results to the Rel-grams data 
(learned from newswire)~\cite{Balasubramanianetal13}.  
For event pairs from films, we randomly sample 
100 high-CP event pairs ensuring that each of the
first events of the pairs are distinct. We use the publicly available
search interface for Rel-grams to find tuples with the same first
event for direct comparison of content of the learned knowledge. 
We set the co-occurrence window to 5, and select the Rel-gram
tuples with the highest \# 50 (FS) (frequency of first statement
occurring before second statement within a window of 50) to choose
high-quality tuples. 
We evaluate the extracted Rel-gram tuples using the same 
Mechanical Turk HIT described above. Table~\ref{tab:film-vs-relgram} shows 
Mechanical Turk evaluation results for our method on 
films vs. Rel-grams: in 81\% questions, humans judge the high-CP 
pairs to be more likely to manifest a causal relation.
We believe this is because the fine-grained 
event pairs we learn do not exist in the Rel-gram collections and thus
the Rel-gram tuples that matched our first events are not highly coherent,
despite the filtering we applied.

\begin{table}[hbt!]
\centering
\small
\begin{tabularx}{2.4in}{p{0.75in} | c | c}
\toprule
{\bf Dataset} & {\bf Film} & {\bf Rel-gram Tuples} \\
\midrule
Percentage of causal relation & 81 \% & 19 \%  \\ 
\bottomrule
\end{tabularx}
\caption{Percentage of pairs judged as causal by AMT workers. Film vs. Rel-Grams. \label{tab:film-vs-relgram}}
\end{table}

For event pairs from camping blogs, we evaluate all 100 high-CP pairs
in a Mechanical Turk study where Turkers are asked to choose whether
an event pair has causal relation or not.
We also evaluate Rel-gram tuples using the same task. However,
Rel-grams are not sorted by topic. To find tuples relevant 
to Camping Trip, we use our top 10 indicative events and extracted 
all the Rel-gram tuples that included at least one event corresponding 
to one of the Camping indicative events, e.g. \emph{go camp}.
We remove any tuple with frequency less than 25 and sort the rest 
by the total symmetrical conditional probability. 
The evaluation results presented in Table~\ref{tab:cam-vs-relgram} 
show that 82\% of the blog paurs were labeled as causal, where as 
only 42\% of the Rel-gram pairs were labeled as causal. 
We argue that this is mainly due to the limitations of the newswire 
data which does not contain the fine-grained everyday events that 
we have extracted from our corpus.

\begin{table}[hbt!]
\centering
\small
\begin{tabularx}{2.6in}{p{0.75in} | c | c}
\toprule
{\bf Dataset} & {\bf Camping} & {\bf Rel-gram Tuples} \\
\midrule
Percentage of causal relation & 82 \% & 42 \%  \\ 
\bottomrule
\end{tabularx}
\caption{Percentage of pairs judged as causal by AMT workers. Camping  blogs vs. Rel-Grams. \label{tab:cam-vs-relgram}}
\end{table}

Next, we compare our results to the event pairs in VerbOcean 
(learned from newswire) with
the {\sc happens-before} relation~\cite{ChklovskiPantel04}. 
We use all 6497 event pairs from VerbOcean, comparing with our
684 event pairs from films and 100 event pairs from camping blogs 
with high CP scores. Our result shows that there are 12 event pairs
that exist in both VerbOcean and films, e.g. \emph{turn - leave} 
and \emph{slow - stop},
and there is only one event pair that exist in both VerbOcean and 
camping blogs: \emph{pack - leave}. This confirms that most causal relations 
learned from other narrative genres do not exist in the currently 
available knowledge bases extracted from newswire. A number 
of event pairs from these collections share the first event, 
e.g. \emph{dig - find} and \emph{scan - spot} from films 
vs. \emph{dig - repair} and \emph{scan - upload} from VerbOcean; 
\emph{drive - park} and \emph{pick - eat} from blogs vs. 
\emph{drive - drag} and \emph{pick - plunk} from VerbOcean.

Finally, we compare our high-CP pairs learned from film to the high-CP 
event pairs from \newcite{BeamerGirju09}, learned from only 173 films.  
There is no public release of Beamer and Girju's event pairs, thus we take
the 29 event pairs with high CP score presented in the paper.
A total of 14 of their 29 pairs are also in our top 684 film pairs. 
These include pairs such as \emph{swerve - avoid}, \emph{leave - stand} 
and \emph{unlock - open}. However on our larger
genre-sorted corpus we also learn pairs such as \emph{grab - haul}, 
\emph{scratch - claw} and \emph{saddle- mount} that do not exist in 
their collection.

\section{Conclusions and Future Work}
\label{sec:conclusion}

Causality is often granular in nature with major events related to the
occurrence of finer-grained events. In this position paper,
we argue that the focus on newswire has inhibited attempts to
learn fine-grained causal relations between everyday events, and that other
narrative genres better support such learning.  We use unsupervised
methods to extract fine-grained causal event relations from films and
blog posts about camping. 

We show that more than 80\% of the relations we learn
are evaluated as causal, and that topical
coherence plays an important role in modeling event
relations. We also show that the causal
knowledge we learn from other narrative genres does not exist in current
event collections induced from newswire.  
We plan to expand our genre-specific experiments on the films corpus in future, as well as using other narrative  datasets, like restaurant reviews, to extract fine-grained causal knowledge about events.




\bibliographystyle{acl_natbib}



\end{document}